\documentclass{article}
\usepackage{spconf,amsmath,graphicx}
\usepackage{booktabs}

\title{Adaptive Bi-directional Attention: Exploring Multi-Granularity Representations for Machine Reading Comprehension}
\name{Nuo Chen$^1$,  Fenglin Liu$^1$, Chenyu You$^2$, Peilin Zhou$^1$,  Yuexian Zou$^{1,3,*}$
\thanks{The research work is supported by Shenzhen basic research program (NO: JCYJ20170817155939233). Special acknowledgements are given to AOTO-PKUSZ JointResearch Center for Artificial Intelligence on Scene Cognition Technology Innovation for its support.}
\thanks{*Corresponding author}}
\address{
  $^1$ADSPLAB, School of ECE, Peking University, Shenzhen, China\\
  $^2$Department of Electrical Engineering, Yale University, CT, USA\\
  $^3$Peng Cheng Laboratory, Shenzhen, China\\
  }
\begin{document}
%
\maketitle
\begin{abstract}
Recently, the attention-enhanced multi-layer encoder, such as Transformer, has been extensively studied in Machine Reading Comprehension (MRC). To predict the answer, it is common practice to employ a predictor to draw information only from the final encoder layer which generates the \textit{coarse-grained} representations of the source sequences, i.e., passage and question.
Previous studies have shown that  the representation of source sequence becomes more \textit{coarse-grained} from \textit{fine-grained} as the encoding layer increases.
It is generally believed that with the growing number of layers in deep neural networks, the encoding process will gather  relevant information for each location increasingly, resulting in more \textit{coarse-grained} representations, which adds the likelihood of similarity to other locations (referring to homogeneity). Such a phenomenon will mislead the model to make wrong judgments so as to degrade the performance.
To this end, we propose a novel approach called Adaptive Bidirectional Attention, which adaptively exploits the source representations of different levels to the predictor. Experimental results on the benchmark dataset, SQuAD 2.0  demonstrate the effectiveness of our approach, and  the results are better than the previous state-of-the-art model by 2.5$\%$ EM and 2.3$\%$ F1 scores.
\end{abstract}
\section{Introduction}
\label{sec:intro}
Machine reading comprehension (MRC) \cite{you2020knowledge,DBLP:conf/iclr/YuDLZ00L18,liu2018stochastic} is a long-standing task that aims to teach the machine how to read and comprehend a given source sequence, i.e., passage/paragraph, then answer its corresponding given questions automatically. It has large amounts of real application scenarios such as question answering and dialog systems.

\begin{figure*}[t]
\centering
\includegraphics[width=1\linewidth]{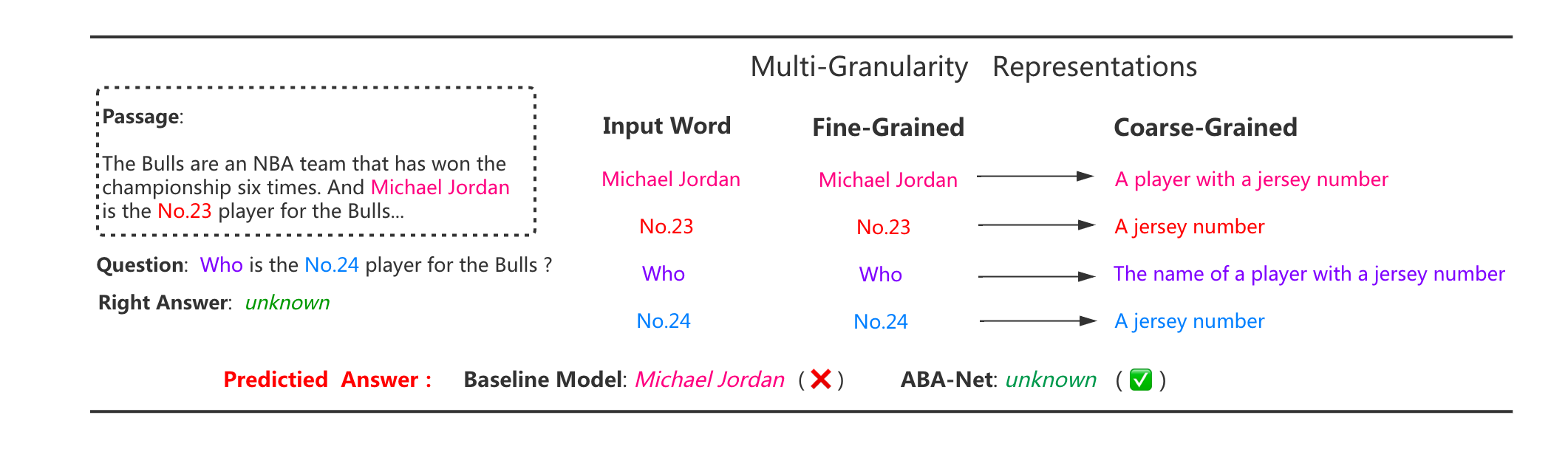}
\caption{An visualization example of a Question-Answer pair for a passage.}
  \label{fig:exg}
\end{figure*}

Recently, the attention-enhanced multi-layer encoder, e.g., Transformer \cite{DBLP:conf/nips/VaswaniSPUJGKP17}, ALBERT \cite{DBLP:conf/iclr/LanCGGSS20}, RoBERTa \cite{DBLP:journals/corr/abs-1907-11692}, and XLNet \cite{yang2019xlnet}, which is based solely on attention mechanisms \cite{Bahdanau2015seq2seq,you2020contextualized} and eliminates recurrence entirely, has been proposed and has established the state-of-the-art in multiple challenging MRC datasets \cite{DBLP:conf/emnlp/RajpurkarZLL16,reddy-etal-2019-coqa,you2020data}. 
However, under the multi-layer deep learning setting, the representations of source sequence will become more \textit{coarse-grained} from \textit{fine-grained} with the growing number of  encoder layers.
Following in \cite{DBLP:conf/iclr/HuangZSC18}, Figure~\ref{fig:exg} illustrates that as the multi-layer encoder processes the source sequences, each input word will gradually gather  related information increasingly as more layers are used, resulting in more abstract representations, i.e., from \textit{fine-grained} to \textit{coarse-grained} representations that adds the likelihood to be similar with other positions (Homogeneous Phenomenon). For those representations output by different encoder layers, the common practice for the current answer predictor is to draw information (\textit{coarse-grained} representations) only from the final encoder layer. However, it should be intuitive that \textit{coarse-grained} representations are good at expressing the overall meaning of the source sequence, but are less precise in the finer details. If we also exploit \textit{fine-grained} representations for the answer predictor, it will help the predictor find  precise source information  
to a large extent and give answers more accurately.
As we observe in Figure~\ref{fig:exg}, due to the baseline model only focuses on the \textit{coarse-grained} representations, it gives an incorrect answer.
In contrast, our method exploits detailed and accurate information, i.e., the \textit{fine-grained} representations of \textit{NO.23} and \textit{NO.24}, resulting in helping the model focus on the correct source information and predict the correct answer.

In this paper, we argue that it would be better if the predictor could exploit representations of different granularity from the encoder, providing different views of the source sequences, such that the expressive power of the model could be fully utilized.
As a consequence, we propose a novel method called   Adaptive Bidirectional Attention to dynamically provide multi-granularity source representations for better predicting the correct answer.
As shown in Figure~\ref{fig:attention}, the proposed approach builds connections with each encoder layer, so that the MRC model not only can exploit the \textit{coarse-grained} representation, of the source sequence, which is instrumental in language modeling, but also has the ability to exploit the \textit{fine-grained} representations of the source sequence, which help predict  more precise answers. Hence, the answer predictor is encouraged to use source representations of different granularity, exploiting the expressive power of the model.  Our experimental results show that  Adaptive Bidirectional Attention has substantial gains in predicting answers correctly.

\begin{figure}[t]
\centering
\includegraphics[width=0.95\linewidth]{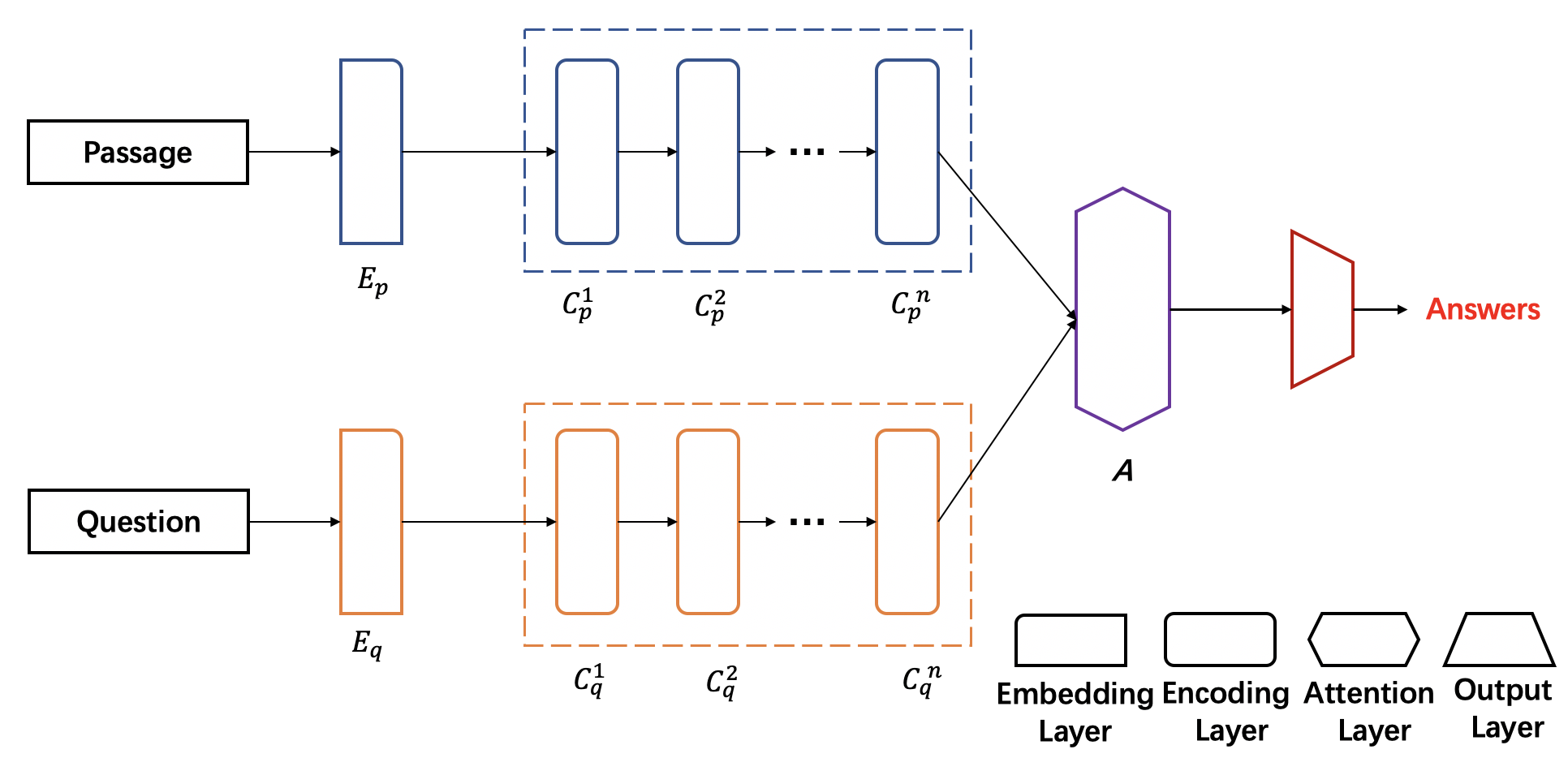}
\caption{A flow diagram of Machine Reading Comprehension architecture. }
  \label{fig:model}
 \end{figure}

\section{Model} 
\subsection{Problem Formulation}
In this paper, the reading comprehension task is defined as follows: Given a passage/paragraph with $l$ words $P=\{p_1,p_2,p_3,...,p_l\}$ and a question with $m$ words  $Q=\{q_1,q_2$ $,q_3,...,q_m\}$ as inputs, the goal is to find an answer span $A$ in $P$. If the question is answerable, the answer span $A$ exists in $P$ as a sequential text string; Otherwise,  $A$ is set to an empty string indicating that the question is unanswerable. Formally, the answer is formulated as $A=\{p_{begin},... ,p_{end}\}$. In the case of unanswerable questions, $A$ denotes the last token of the passage.
\subsection{Model Overview}
Generally, typical MRC models contain four major parts: an embedding layer, several encoding layers, an attention layer  and an output layer, as shown in Figure~\ref{fig:model}. $Embedding$ $Layer$ is responsible for encoding each token in the passage and question into a fixed-length vector. In the \textit{Encoding Layers}, the models move forward to extract contextual cues from the surrounding words and sentences to form higher level granularity representations with some efficient methods such as transformers, BiLSTMs \cite{hochreiter1997long} and CNNs, to name a few. Next, \textit{Attention Layer} plays a key role in MRC development, which has the ability of extracting relationship between words at various positions of context and question, and even focuses on the most important part of the interaction. Finally, the \textit{Output Layer} decodes to the probability that each position is the beginning or end of the answer span so as to predict an answer.

\subsection{Adaptive Bidirectional Attention}
  
 \begin{figure*}[t]
\centering
\includegraphics[width=1\linewidth]{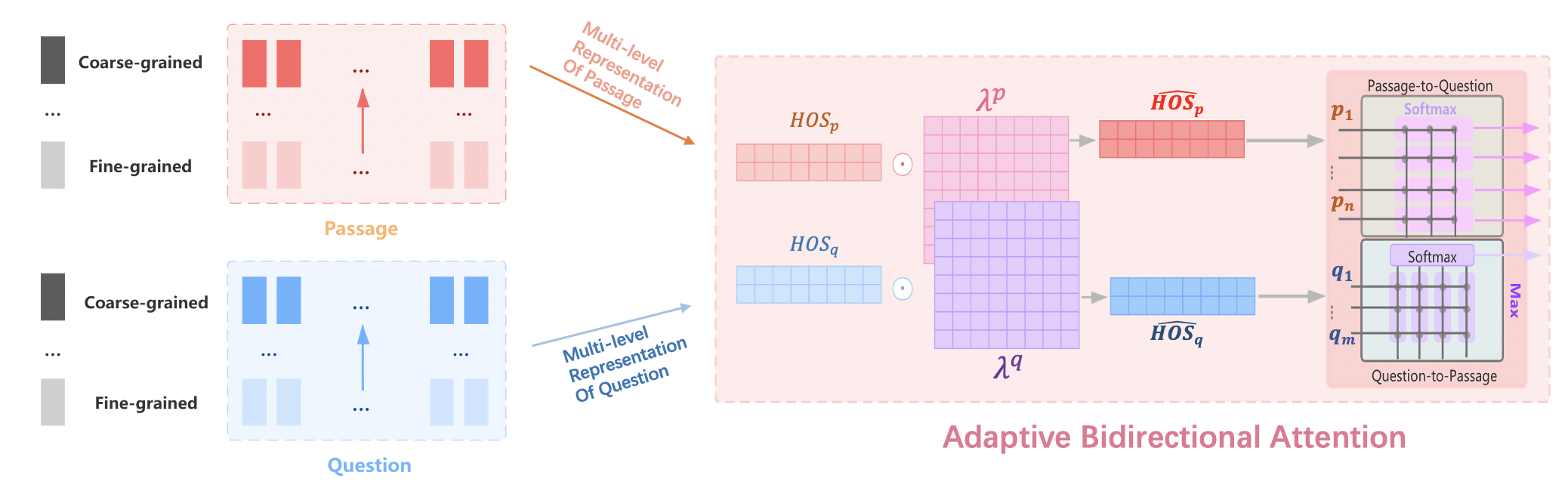}
\caption{An overview  of Adaptive Bidirectional Attention. $\odot$ is the element-wise multiplication.} \label{fig:attention} 
\end{figure*}
As illustrated in Section \ref{sec:intro}, we propose Adaptive Bidirectional Attention to encourage the answer predictor (output layer) to take full advantage of the expressive power of the multi-layer encoders through exploring and exploiting the multi-granularity representations of source sequences.

To simplify notation, we first define the output of Embedding layer are $E_p$ for passage and $E_q$ for question, separately. Then the output of Encoding layers are formulated as  $C_p^1$, $C_p^2$, ..., $C_p^n$; $C_q^1$, $C_q^2$, ..., $C_q^n$, sequentially. Next, the output of the Attention layer is  $\textbf{A}$. After multiple layers  extracting different levels of presentations  of each word, we conduct Adaptive Bidirectional Attention from question to passage and passage to question based on generated representations to take advantage of all layers of representation reasonably, which is showed in Figure~\ref{fig:attention}. We define a novel concept -- 'history of semantic', which denotes multi-granularity representations extracted by the model before the current layer. Thus history-of-semantic vectors can be defined as:
 \begin{equation}
HOS_p=[E_p;C_p^{1};C_p^{2};...;C_p^{n};\textbf{A}] 
\end{equation}
\begin{equation}
HOS_q=[E_q;C_q^{1};C_q^{2};...;C_q^{n};\textbf{A}] 
\end{equation}
 
 \textbf{Adaptive.}\quad In order to obtain multi-level representations   connection between the encoder layers. We have designed the following function:
\begin{equation}
 \lambda ^{p} \odot{HOS_p^{T}}=\widehat{HOS_p^{T}}  \quad \lambda ^{q} \odot{HOS_q^{T}}=\widehat{HOS_q^{T}}	
 \end{equation}
 where the matrix $\lambda$  is  trainable. Notice that at the beginning of training, in order to retain the original semantics of each layer, we will initialize the first column of this matrix to all 1, the remaining columns are all 0.

\textbf{Bidirectional  Attention.} \quad In this component,  we are aiming to dynamically compute attention of the embedding vectors from previous layers each time, as well as the multi-granularity  representations generated from the different layers are allowed to flow through to the downstream output layer. Like most high performing models, such as  \cite{DBLP:journals/corr/XiongZS16,DBLP:conf/iclr/SeoKFH17}, we construct passage-to-question and question-to-passage attention, respectively. The attention function in  \cite{DBLP:conf/iclr/SeoKFH17} is used to compute the similarity score between passages and questions. First, we calculate the similarities between each pair words in passages and questions, and render a similarity matrix $H\in \textbf{R}^{n\times{m}}$. H is computed as:
     \begin{equation}H=dropout(f_{attention}[\widehat{HOS_p},\widehat{HOS_q}])\end{equation}
  After that, we use the strategy of  \cite{DBLP:conf/iclr/YuDLZ00L18} to normalize each row of H by applying the softmax function, and then get a matrix $ \hat H$. Then the passage-to-question attention is computed as $M= \hat H \cdot (\widehat{HOS_q})^T $. Following DCN \cite{DBLP:conf/iclr/XiongZS17} , we compute the column normalized matrix $\bar{H}$ of H by softmax function, and the question-to-passage attention is $S = \hat H \cdot \bar H^T \cdot (\widehat{HOS_p})^T$. At last, our method use a simple concatenation as following to get the final representation, which shows good performances in our experiments: $I=[\widehat{HOS_p};M;\widehat{HOS_p} \odot M;\widehat{HOS_p} \odot S]$. In this work,  selected MRC models will take $I$ as input of the \textit{Output Layer}.

\begin{table}[]
\caption{Comparison study of Adaptive Bidirectional Attention on SQuAD 2.0. }
   \centering
        \begin{tabular}{l c c}
       \toprule
         \textbf{Base Model} & EM&F1 \\
       \toprule
    BiDAF++ \cite{DBLP:conf/iclr/ShenZL0Z18}& 65.6&68.7 \\
    \ + \textbf{Adaptive Bidirectional Attention} & \bfseries{67.2}&\bfseries{70.8} \\ \hline
    QANet \cite{DBLP:conf/iclr/YuDLZ00L18}& 65.4&67.2 \\
    \ +  \textbf{Adaptive Bidirectional Attention} & \bfseries {66.9}&\bfseries{69.1} \\ \hline
    SAN \cite{liu2018stochastic} & 68.6&71.4 \\  
    \ + \textbf{Adaptive Bidirectional Attention}& \bfseries {70.1}&\bfseries{73.7} \\ \hline
    SDNet \cite{DBLP:journals/corr/abs-1812-03593}& 76.7&79.8 \\
    \ + \textbf{Adaptive Bidirectional Attention} & \bfseries{78.5}&\bfseries{81.7} \\ \hline
    SGNet\cite{DBLP:conf/aaai/0001WZDZ020} & 85.1&87.9 \\
    \ + \textbf{Adaptive Bidirectional Attention} & \bfseries{87.6}&\bfseries{90.2} \\  

     \toprule
    \end{tabular}
    
    \label{tab:my_label_3}
\end{table}

\section{Experiments }
In this section, we conduct a series of experiments to study the performance of our method. We will primarily benchmark our method on the SQuAD 2.0 dataset \cite{DBLP:conf/acl/RajpurkarJL18}.
\begin{figure*}[t]
\centering
\includegraphics[width=0.85\linewidth]{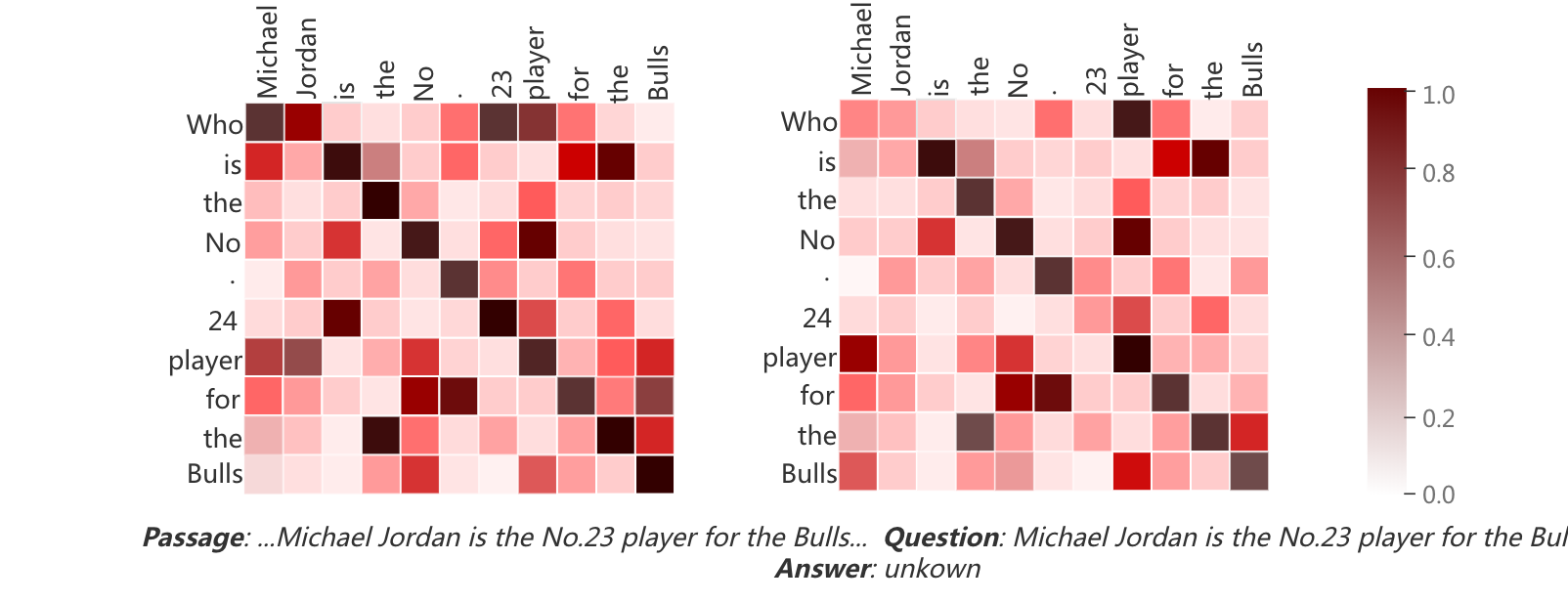}
\caption{
Visualization of Syntax-guided Attention (left) and Adaptive Bidirectional Attention (right). The example is the same as the showed in Figure \label{fig:example}.  
For Syntax-guided attention, $who$ focuses on $Michael$ and \textit{\textbf{23}}, and there is a great similarity between \textit{\textbf{23}} and \textit{\textbf{24}}, which would mislead the model to make the wrong judgment. In contrast, this phenomenon does not  happen in our Adaptive Bidirectional attention. } \label{fig:attention1}
\end{figure*}
\subsection{Dataset And Experimental Settings}
\subsubsection{Dataset}  
We  evaluate our method on SQuAD 2.0 dataset \cite{DBLP:conf/acl/RajpurkarJL18}, a new MRC dataset which is a combination of SQuAD 1.0 and additional unanswerable question-answer pairs. Specifically, the number of questions that can be answered is about 100K; while the unanswerable questions are around 53K. For the SQuAD 2.0 challenge, MRC models must answer questions not only where possible, but also when the paragraph does not support any answers.   
 


\subsubsection{Baseline Models}
To prove the generality and effectiveness of our proposed method, we choose several previous state-of-the-art QA models. In a nutshell, the architectures of  QA models are briefly summarized below.  BiDAF++ \cite{DBLP:conf/iclr/ShenZL0Z18} is a  very representative work for machine comprehension, which proposes a Bi-Directional Attention Flow to obtain query-aware contextual representations.  QANet \cite{DBLP:conf/iclr/YuDLZ00L18}, an  atypical  but  very  effective  QA model,  which  completely  contains  no  recurrent  networks and consists of stacked encoder blocks with convolution networks  and  self-attention.  SAN \cite{liu2018stochastic} is  an alternative multi-step reasoning neural network for MRC, simple yet robust.   SDNet\cite{DBLP:journals/corr/abs-1812-03593} and SGNet \cite{DBLP:conf/aaai/0001WZDZ020} both can be viewed as BERT-based models, which take BERT as their backbone encoders.
\subsection{Results }
In this study, we use F1 and EM as evaluation metrics. F1 measures the part of the overlapping mark between the predicted answer and the  ground-truth answer, and if the prediction is exactly the same as the ground truth, the exact match (EM) score is 1, otherwise it is 0. Specifically, we add Adaptive Bidirectional Attention on some end-to-end MRC models to compare with their initial version (i.e. base models in Table~\ref{tab:my_label_3}) on SQuAD 2.0. As can be seen from Table~\ref{tab:my_label_3}, after adding Adaptive Bidirectional Attention, the performance of these models could be improved to varying degrees. As for SGNet\cite{DBLP:conf/aaai/0001WZDZ020}, our method even improves EM and F1 scores by 2.5$\%$ and 2.3$\%$, separately.
This also proves the versatility of this attention mechanism.

\section{Visualization} 
In this component, to have an insight that how Adaptive Bidirectional Attention works, we draw attention distributions of the Syntax-guidedAttention of SGNet \cite{DBLP:conf/aaai/0001WZDZ020} and our proposed Adaptive Bidirectional attention, as shown in Figure~\ref{fig:attention1}. The visualization verifies that benefiting from Adaptive Bidirectional Attention, our method is effective at  distinguishing and utilizing different layers of presentation of each word in the context and the question, guiding the downstream layer to collect more relevant semantics to make predictions so as to help MRC models in predicting the answer more accurately.

\section{Conclusion}
In this paper, we propose a novel attention mechanism Adaptive Bidirectional Attention to explore and exploit  multi-granularity representations of source sequences for machine reading comprehension. In particular, our method can adaptively exploit the source representations of different levels to the predictor.  More concretely, Adaptive Bidirectional Attention is used to guide attention learning and  reasonably leverages source representations of different levels for question answering by dynamically capturing the connection between all layers of representations of each word.  The experimental results show the effectiveness of our approach on the public benchmark dataset, SQuAD 2.0.  Future work can include an extension of employing Adaptive Bidirectional Attention to other natural language processing tasks.



\vfill\pagebreak

\bibliographystyle{IEEE.bst}
\bibliography{main.bib}

\begin{thebibliography}{10}

\bibitem{you2020knowledge}
Chenyu You, Nuo Chen, and Yuexian Zou,
\newblock ``Knowledge distillation for improved accuracy in spoken question
  answering,''
\newblock {\em arXiv preprint arXiv:2010.11067}, 2020.

\bibitem{DBLP:conf/iclr/YuDLZ00L18}
Adams~Wei Yu, David Dohan, Minh{-}Thang Luong, Rui Zhao, Kai Chen, Mohammad
  Norouzi, and Quoc~V. Le,
\newblock ``Qanet: Combining local convolution with global self-attention for
  reading comprehension,''
\newblock in {\em {ICLR} (Poster)}. 2018, OpenReview.net.

\bibitem{liu2018stochastic}
Xiaodong Liu, Wei Li, Yuwei Fang, Aerin Kim, Kevin Duh, and Jianfeng Gao,
\newblock ``Stochastic answer networks for squad 2.0,''
\newblock {\em arXiv preprint arXiv:1809.09194}, 2018.

\bibitem{DBLP:conf/nips/VaswaniSPUJGKP17}
Ashish Vaswani, Noam Shazeer, Niki Parmar, Jakob Uszkoreit, Llion Jones,
  Aidan~N. Gomez, Lukasz Kaiser, and Illia Polosukhin,
\newblock ``Attention is all you need,''
\newblock in {\em {NIPS}}, 2017, pp. 5998--6008.

\bibitem{DBLP:conf/iclr/LanCGGSS20}
Zhenzhong Lan, Mingda Chen, Sebastian Goodman, Kevin Gimpel, Piyush Sharma, and
  Radu Soricut,
\newblock ``{ALBERT:} {A} lite {BERT} for self-supervised learning of language
  representations,''
\newblock in {\em {ICLR}}. 2020, OpenReview.net.

\bibitem{DBLP:journals/corr/abs-1907-11692}
Yinhan Liu, Myle Ott, Naman Goyal, Jingfei Du, Mandar Joshi, Danqi Chen, Omer
  Levy, Mike Lewis, Luke Zettlemoyer, and Veselin Stoyanov,
\newblock ``Roberta: {A} robustly optimized {BERT} pretraining approach,''
\newblock {\em CoRR}, vol. abs/1907.11692, 2019.

\bibitem{yang2019xlnet}
Zhilin Yang, Zihang Dai, Yiming Yang, Jaime Carbonell, Russ~R Salakhutdinov,
  and Quoc~V Le,
\newblock ``Xlnet: Generalized autoregressive pretraining for language
  understanding,''
\newblock in {\em Advances in neural information processing systems}, 2019, pp.
  5754--5764.

\bibitem{Bahdanau2015seq2seq}
Dzmitry Bahdanau, Kyunghyun Cho, and Yoshua Bengio,
\newblock ``Neural machine translation by jointly learning to align and
  translate,''
\newblock in {\em {ICLR}}, 2015.

\bibitem{you2020contextualized}
Chenyu You, Nuo Chen, and Yuexian Zou,
\newblock ``Contextualized attention-based knowledge transfer for spoken
  conversational question answering,''
\newblock {\em arXiv preprint arXiv:2010.11066}, 2020.

\bibitem{DBLP:conf/emnlp/RajpurkarZLL16}
Pranav Rajpurkar, Jian Zhang, Konstantin Lopyrev, and Percy Liang,
\newblock ``Squad: 100, 000+ questions for machine comprehension of text,''
\newblock in {\em {EMNLP}}. 2016, pp. 2383--2392, The Association for
  Computational Linguistics.

\bibitem{reddy-etal-2019-coqa}
Siva Reddy, Danqi Chen, and Christopher~D. Manning,
\newblock ``{C}o{QA}: A conversational question answering challenge,''
\newblock {\em Transactions of the Association for Computational Linguistics},
  vol. 7, Mar. 2019.

\bibitem{you2020data}
Chenyu You, Nuo Chen, Fenglin Liu, Dongchao Yang, and Yuexian Zou,
\newblock ``Towards data distillation for end-to-end spoken conversational
  question answering,''
\newblock {\em arXiv preprint arXiv:2010.08923}, 2020.

\bibitem{DBLP:conf/iclr/HuangZSC18}
Hsin{-}Yuan Huang, Chenguang Zhu, Yelong Shen, and Weizhu Chen,
\newblock ``Fusionnet: Fusing via fully-aware attention with application to
  machine comprehension,''
\newblock in {\em 6th International Conference on Learning Representations,
  {ICLR} 2018, Vancouver, BC, Canada, April 30 - May 3, 2018, Conference Track
  Proceedings}. 2018, OpenReview.net.

\bibitem{hochreiter1997long}
Sepp Hochreiter and J{\"u}rgen Schmidhuber,
\newblock ``Long short-term memory,''
\newblock {\em Neural computation}, vol. 9, no. 8, pp. 1735--1780, 1997.

\bibitem{DBLP:journals/corr/XiongZS16}
Caiming Xiong, Victor Zhong, and Richard Socher,
\newblock ``Dynamic coattention networks for question answering,''
\newblock {\em CoRR}, vol. abs/1611.01604, 2016.

\bibitem{DBLP:conf/iclr/SeoKFH17}
Min~Joon Seo, Aniruddha Kembhavi, Ali Farhadi, and Hannaneh Hajishirzi,
\newblock ``Bidirectional attention flow for machine comprehension,''
\newblock in {\em {ICLR} (Poster)}. 2017, OpenReview.net.

\bibitem{DBLP:conf/iclr/XiongZS17}
Caiming Xiong, Victor Zhong, and Richard Socher,
\newblock ``Dynamic coattention networks for question answering,''
\newblock in {\em 5th International Conference on Learning Representations,
  {ICLR} 2017, Toulon, France, April 24-26, 2017, Conference Track
  Proceedings}. 2017, OpenReview.net.

\bibitem{DBLP:conf/iclr/ShenZL0Z18}
Tao Shen, Tianyi Zhou, Guodong Long, Jing Jiang, and Chengqi Zhang,
\newblock ``Bi-directional block self-attention for fast and memory-efficient
  sequence modeling,''
\newblock in {\em 6th International Conference on Learning Representations,
  {ICLR} 2018, Vancouver, BC, Canada, April 30 - May 3, 2018, Conference Track
  Proceedings}. 2018, OpenReview.net.

\bibitem{DBLP:journals/corr/abs-1812-03593}
Chenguang Zhu, Michael Zeng, and Xuedong Huang,
\newblock ``Sdnet: Contextualized attention-based deep network for
  conversational question answering,''
\newblock {\em CoRR}, vol. abs/1812.03593, 2018.

\bibitem{DBLP:conf/aaai/0001WZDZ020}
Zhuosheng Zhang, Yuwei Wu, Junru Zhou, Sufeng Duan, Hai Zhao, and Rui Wang,
\newblock ``Sg-net: Syntax-guided machine reading comprehension,''
\newblock in {\em {AAAI}}. 2020, pp. 9636--9643, {AAAI} Press.

\bibitem{DBLP:conf/acl/RajpurkarJL18}
Pranav Rajpurkar, Robin Jia, and Percy Liang,
\newblock ``Know what you don't know: Unanswerable questions for squad,''
\newblock in {\em {ACL} {(2)}}. 2018, pp. 784--789, Association for
  Computational Linguistics.

\end{thebibliography}

\end{document}